\pdfoutput=1

\documentclass[11pt]{article}

\usepackage[final]{acl}

\usepackage{times}
\usepackage{latexsym}

\usepackage[T1]{fontenc}

\usepackage[utf8]{inputenc}

\usepackage{microtype}

\usepackage{inconsolata}

\usepackage{graphicx}

\usepackage{comment}
\usepackage{booktabs}
\usepackage{multirow}
\usepackage{combelow}
\usepackage{tabularx}
\usepackage{xcolor}
\usepackage{lipsum}
\usepackage{pifont}
\usepackage{amsmath}
\usepackage{authblk}

\newcommand{\xmark}{\ding{55}}%

\title{MoRoVoc: A Large Dataset for Geographical Variation Identification of the Spoken Romanian Language
}

\author{\textbf{Andrei-Marius Avram$^{1}$, Ema-Ioana B\u{a}nescu$^{1}$}, \\ 
\textbf{Anda-Teodora Robea$^{1}$, Dumitru-Clementin Cercel$^{1}$\thanks{Corresponding author.}, Mihaela-Claudia Cercel$^{2,3}$}  \\
$^{1}$National University of Science and Technology POLITEHNICA Bucharest, Romania \\
$^{2}$Paris 1 Panthéon-Sorbonne University, Paris, France \\
$^{3}$University of Bucharest, Bucharest, Romania \\
\tt {\{andrei\_marius.avram,ema\_ioana.banescu,anda\_teodora.robea\}@stud.acs.upb.ro,} \\ 
\tt {dumitru.cercel@upb.ro, mihaela-claudia.cercel@drept.unibuc.ro}
\\
}

\begin{document}
\maketitle
\begin{abstract}

This paper introduces MoRoVoc, the largest dataset for analyzing the regional variation of spoken Romanian. It has more than 93 hours of audio and 88,192 audio samples, balanced between the Romanian language spoken in Romania and the Republic of Moldova. We further propose a multi-target adversarial training framework for speech models that incorporates demographic attributes (i.e., age and gender of the speakers) as adversarial targets, making models discriminative for primary tasks while remaining invariant to secondary attributes. The adversarial coefficients are dynamically adjusted via meta-learning to optimize performance. Our approach yields notable gains: Wav2Vec2-Base achieves 78.21\% accuracy for the variation identification of spoken Romanian using gender as an adversarial target, while Wav2Vec2-Large reaches 93.08\% accuracy for gender classification when employing both dialect and age as adversarial objectives. 

\end{abstract}

\section{Introduction}

Dialect identification poses a fundamental challenge in speech processing, and low-resource languages face persistent difficulties due to the scarcity of annotated corpora. Despite its millions native speakers, Romanian remains underrepresented in speech technology research.
 
We introduce MoRoVoc, the largest corpus for Romanian spoken dialect identification to date, containing 93+ hours of annotated speech with 88,192 audio samples balanced between standard Romanian and Moldavian dialects. The dataset includes comprehensive demographic metadata and is derived from high-quality parliamentary recordings.

Our methodological contribution introduces a novel multi-target adversarial training framework for fine-tuning speech models through gradient reversal \cite{avram2024histnero,avram2025rolargesum}. This approach incorporates multiple demographic attributes as adversarial targets \cite{zhao2022adversarial} with dynamically adjusted adversarial coefficients through meta-learning \cite{vettoruzzo2024advances}, allowing models to learn representations that are discriminative for the primary task while remaining invariant with secondary attributes.

Our experiments with Wav2Vec2 \cite{NEURIPS2020_92d1e1eb} demonstrate that our multi-target adversarial training significantly improves performance across all classification tasks. For spoken dialect identification, Wav2Vec2-Base achieves 78.21\% accuracy (a 5.20\% improvement over baseline) when using gender as adversarial target, while Wav2Vec2-Large reaches 77.53\% accuracy when both gender and age serve as adversarial targets.

The main contributions of our work can be summarized as follows:
\begin{itemize}
    \item We present MoRoVoc, the largest corpus for Romanian spoken dialect identification, which includes detailed gender and age annotations, and is available for public use\footnote{\url{https://huggingface.co/datasets/avramandrei/morovoc}}.
    \item We conducted an in-depth analysis of MoRoVoc and compared it with another dataset of spoken dialect identification available in the literature (see the Appendix \ref{app:related_work}).
    \item We propose a novel multi-target adversarial training framework for speech models that employs gradient reversal on multiple attributes, adapting their coefficients using meta-learning, which results in an improvement of up to 5.20\% for dialect identification.
\end{itemize}

\section{MoRoVoc Dataset}

\subsection{Data Collection and Annotation}

The audio samples in the MoRoVoc dataset were obtained from publicly available recordings of parliamentary debates from Romania and the Republic of Moldova. These recordings, sourced from official legislative sessions, provide a natural setting in which the speakers exhibit both the standard Romanian accent and distinct Moldavian phonetic traits. To ensure a high-quality and representative corpus for dialect identification, we used a rigorous selection process. 

Initially, each recording was manually segmented to isolate single-speaker segments. The segments containing overlapping speech or exhibiting low intelligibility were then discarded. To ensure reliable dialect labels, all segments were subsequently submitted to local annotators, native speakers proficient in the standard Romanian language, for manual validation. Samples with ambiguous dialect cues were removed from the final collection. 

After curation, the MoRoVoc dataset comprises 88,192 audio samples (i.e., 43,186 for Romanian and 45,006 for Moldavian) with durations ranging from 0.4 to 30 seconds, all recorded at a sampling rate of 22,050 Hz. To prevent models from overfitting to speaker-specific characteristics unrelated to dialect, the dataset was divided into disjoint training, validation, and testing splits, which tried to maintain the original distribution of the data. In total, 77,638 samples were allocated for training, 5,349 for validation, and 5,207 for testing, ensuring that no speaker appears in more than one split.

\subsection{Dataset Statistics}

\begin{table}
    \centering
     \resizebox{0.475\textwidth}{!}{
    \begin{tabular}{|l|c|c|c|c|}
         \toprule
    \textbf{Dialect} & \textbf{Train/Val/Test} & \textbf{\# hours} & \textbf{SNR} & \textbf{SRR} \\
    \midrule
    \multicolumn{5}{|c|}{\textbf{\textit{RoDia}}} \\
    \midrule
    Transylvanian & 427/-/119 & 0.39 & 29.18 & 36.88 \\
    Banatian & 424/-/99 & 0.37 & 23.47 & 35.18 \\
    Moldavian & 384/-/206 & 0.42 & 25.68 & 30.78 \\
    Wallachian & 603/-/106 & 0.51 & 28.67 & 35.62 \\
    Oltenian & 326/-/74 & 0.29 & 26.58 & 31.66 \\
    \midrule
    \textbf{Overall} & \textbf{2164/-/604} & \textbf{1.99} & \textbf{26.89} & \textbf{34.15} \\
    \midrule
    \multicolumn{5}{|c|}{\textbf{\textit{MoRoVoc}}} \\
    \midrule
    Moldavian & 39.6k/2.7k/2.6k & 46.07 & 21.01 & 23.22 \\
    Standard Romanian & 37.9k/2.6k/2.6k & 47.40 & 20.71 & 23.25 \\
    \midrule
    \textbf{Overall} & \textbf{77.6k/5.3k/5.2k} & \textbf{93.48} & \textbf{20.86} & \textbf{23.24} \\
    \bottomrule
    \end{tabular}
    }
    \caption{Overview of dataset statistics comparing MoRoVoc and RoDia. The table details dialect-specific train/validation/test splits, total duration in hours, and key audio quality metrics (SNR and SRR).}
    \label{tab:dataset_stats}
\end{table}

A detailed overview of the audio quality and data distribution statistics for MoRoVoc, compared to RoDia \cite{codrut2024rodia}, is provided in Table~\ref{tab:dataset_stats}. In particular, the table reports the number of samples per dialect (along with their train/validation/test splits), the total duration in hours (93.48 hours), as well as quality metrics including the signal-to-noise ratio (SNR) and the signal-to-reverberation ratio (SRR). Both quality metrics indicate relatively low levels of background noise and reverberation in the recordings.

\begin{figure}
    \centering
    \includegraphics[width=0.75\linewidth]{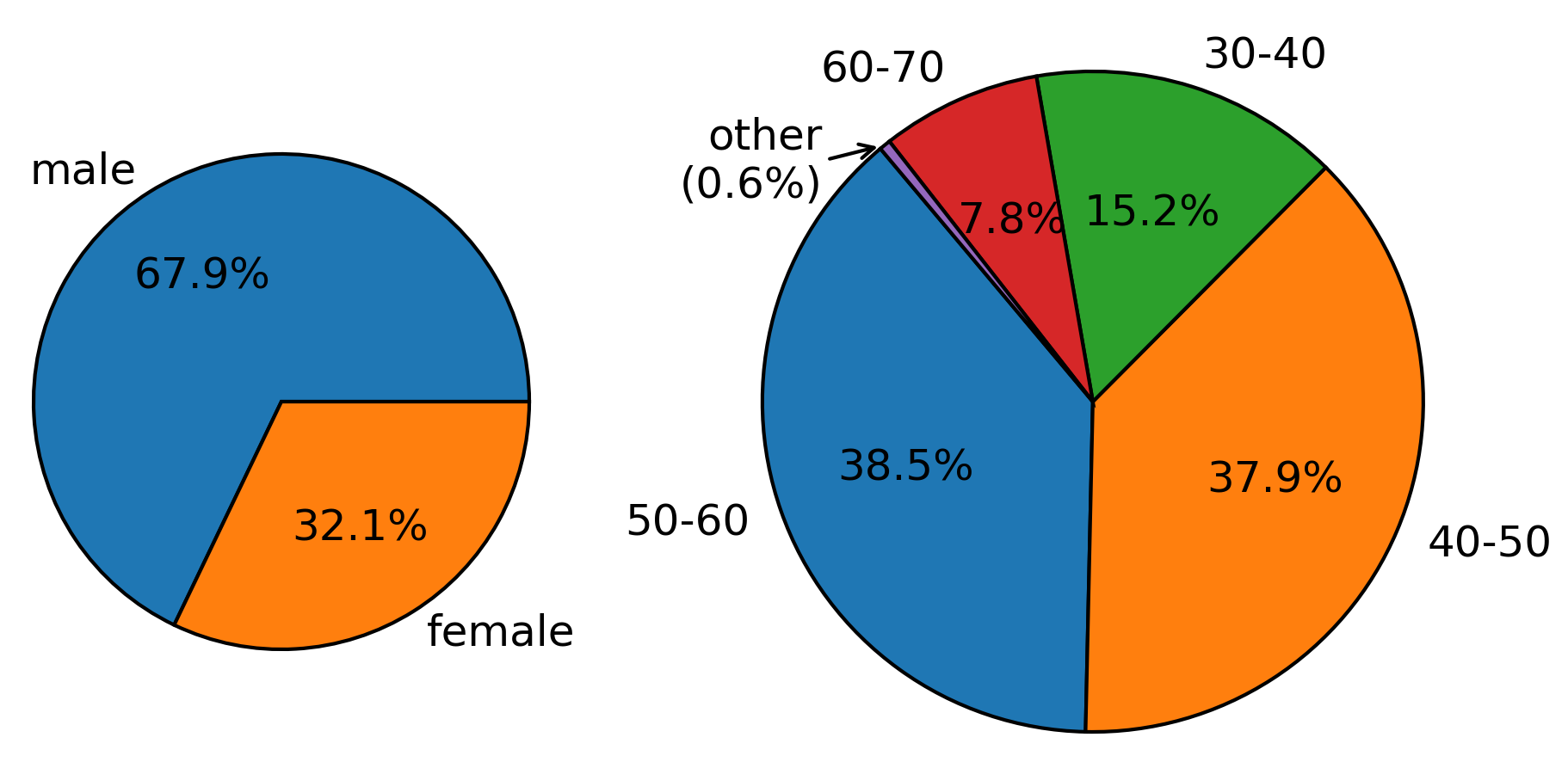}
    \caption{Distribution of gender (left) and age groups (right) in the MoRoVoc dataset.}
    \label{fig:demographics}
\end{figure}

Beyond spoken dialect identification, our annotators also provided demographic labels based on age and gender. Figure~\ref{fig:demographics} illustrates the resulting distributions: 67.9\% of the speakers are male and 32.1\% are female, with the largest age segments being 50–60 (38.5\%) and 40–50 (37.9\%). Younger speakers, aged 30–40, represent 15.2\% of the dataset, while 7.8\% of the speakers are in the 60–70 age range, and 0.6\% fall into the “other” category (i.e., those under 30 years and older than 70 years). This age distribution reflects the typical demographic makeup of parliamentary sessions, where middle-aged and older speakers are more prevalent compared to younger speakers.

Finally, to assess the reliability of these demographic annotations, we measured the inter-rater agreement using the Quadratic Weighted Kappa \cite{vanbelle2016new}, which yielded a value of 0.87. This result indicates a high level of consistency among the annotators. Furthermore, the average accuracy among the annotators was 91\%, which further highlights the strong consensus in labeling the age and gender categories. Textual statistics are presented in Appendix \ref{app:text_stats}.

\section{Methodology}

We introduce a novel fine-tuning strategy for pre-trained Wav2Vec2 models incorporating multiple adversarial targets. Our approach ensures invariance to demographic attributes, including dialect, gender, and age. We consider three configurations, each with a different main task and the remaining attributes as adversarial objectives: 1) \textit{spoken dialect identification} as the main task with gender and age as adversarial goals, 2) \textit{gender classification} as the primary objective with dialect and age as adversarial tasks, and 3) \textit{age classification} with dialect and gender as adversarial tasks.

\subsection{Fine-tuning Wav2Vec2}

We adapt the Wav2Vec2 model by attaching three task-specific classification heads to the shared encoder. In a given configuration, one head is designated as the main task head with the loss $\mathcal{L}_{\text{task}}$, while the remaining two heads generate adversarial losses $\mathcal{L}_{\text{adv}}^{(1)}$ and $\mathcal{L}_{\text{adv}}^{(2)}$.

For each task-specific head, we obtain a global representation of the input speech by averaging the encoder outputs:
\begin{equation}
    \bar{e} = \frac{1}{T} \sum_{t=1}^{T} e_t.
\end{equation}
The prediction is then computed via a linear transformation:
\begin{equation}
    \hat{y} = \sigma(\theta \bar{e} + b),
    \label{eq:task_prediction}
\end{equation}
where $\theta$ and $b$ are the weights and bias, and $\sigma$ denotes an appropriate activation function.

\subsection{Multi-Target Adversarial Training}

The key novelty of our approach is the simultaneous use of multiple adversarial objectives during fine-tuning. The overall training objective is as follows:
\begin{equation}
    \mathcal{L} = \mathcal{L}_{\text{task}} + \lambda_1\, \mathcal{L}_{\text{adv}}^{(1)} + \lambda_2\, \mathcal{L}_{\text{adv}}^{(2)},
    \label{eq:overall_loss}
\end{equation}
where $\lambda_1$ and $\lambda_2$ control the influence of the adversarial losses.

We reverse the gradients from adversarial losses during backpropagation. Let $\theta_{\text{enc}}$ denote the parameters of the Wav2Vec2 encoder. The update is as follows:
\begin{equation}
\resizebox{0.42\textwidth}{!}{$
    \theta_{\text{enc}} \leftarrow \theta_{\text{enc}} - \alpha \left( \nabla_{\theta_{\text{enc}}} \mathcal{L}_{\text{task}} - \gamma_1\, \nabla_{\theta_{\text{enc}}} \mathcal{L}_{\text{adv}}^{(1)} - \gamma_2\, \nabla_{\theta_{\text{enc}}} \mathcal{L}_{\text{adv}}^{(2)} \right),
    $}
    \label{eq:adv_update}
\end{equation}
where $\alpha$ is the learning rate, while $\gamma_1$ and $\gamma_2$ control the strength of the adversarial objectives. This update can be viewed as minimizing:
\begin{equation}
    \mathcal{L}' = \mathcal{L}_{\text{task}} - \lambda_1\, \mathcal{L}_{\text{adv}}^{(1)} - \lambda_2\, \mathcal{L}_{\text{adv}}^{(2)}.
    \label{eq:loss_reversal}
\end{equation}

\subsection{Meta-Learning for Adversarial Coefficient Adaptation}

Selecting appropriate values for coefficients $\gamma_1$ and $\gamma_2$ is challenging. We employ a gradient-based meta-learning strategy to adaptively tune them during training, adopting the model-agnostic meta-learning technique
\cite{finn2017model}.

Thus, after updating model parameters via the inner loop (Equation \ref{eq:overall_loss}), we define a meta-loss on a validation set:
\begin{equation}
    \mathcal{L}_{\text{meta}}(\gamma) = \mathcal{L}_{\text{task}}(x_{\text{val}}, y_{\text{val}}; \theta').
\end{equation}
The outer loop updates the adversarial coefficients:
\begin{equation}
    \gamma \leftarrow \gamma - \eta\, \nabla_\gamma \mathcal{L}_{\text{meta}}(\gamma),
\end{equation}
where $\gamma = (\gamma_1, \gamma_2)$ and $\eta$ is the meta-learning rate. This update optimizes the main task performance on unseen data.

\section{Results}

Table~\ref{tab:results} presents the performance of the Wav2Vec2-Base and Wav2Vec2-Large models on the MoRoVoc dataset\footnote{Further details about the performance of other models with our methodology are given in Appendix \ref{app:additional_models}.}. We evaluated three main tasks while exploring different adversarial objectives. We measure the accuracy, precision, recall, and the F1-score on each task.

\begin{table*}[!ht]
    \centering
    \caption{Wav2Vec2-Base and Wav2Vec2-Large results on MoRoVoc. Up arrow ($\uparrow$) depicts the target optimization objective, down arrow ($\downarrow$) depicts the adversarial optimization objective(s), while \xmark\, outlines that we did not use that optimization objective.}
    \begin{tabular}{|l|wc{3.2em}|wc{3.2em}|wc{3.2em}|c|c|c|c|}
    \toprule
    \textbf{Model} & \textbf{Dialect} & \textbf{Gender} & \textbf{Age} & \textbf{Acc.} & \textbf{P} & \textbf{R} & \textbf{F1} \\
    \midrule
    Wav2Vec2-Base & $\uparrow$ & \xmark & \xmark & 73.01 & 75.08 & 72.02 & 71.80 \\
    Wav2Vec2-Base & $\uparrow$ & $\downarrow$ & \xmark & \textbf{78.21} & \textbf{78.23} & \textbf{77.98} & \textbf{78.05} \\
    Wav2Vec2-Base & $\uparrow$ & \xmark & $\downarrow$ & 76.36 & 76.42 & 76.07 & 76.15 \\
    Wav2Vec2-Base & $\uparrow$ & $\downarrow$ & $\downarrow$ & 70.54 & 72.03 & 71.22 & 70.40 \\
    \midrule
    Wav2Vec2-Base & \xmark & $\uparrow$ & \xmark & 89.60 & 89.93 & 84.34 & 87.04 \\
    Wav2Vec2-Base & $\downarrow$ & $\uparrow$ & \xmark & 89.80 & \textbf{89.90} & 86.63 & 88.23 \\
    Wav2Vec2-Base & \xmark & $\uparrow$ & $\downarrow$ & 88.78 & 89.25 & 84.71 & 86.92 \\
    Wav2Vec2-Base & $\downarrow$ & $\uparrow$ & $\downarrow$ & \textbf{90.21} & 89.89 & \textbf{86.94} & \textbf{88.90} \\
    \midrule
    Wav2Vec2-Base & \xmark & \xmark & $\uparrow$ & 47.32 & \textbf{26.77} & 28.66 & 26.27 \\
    Wav2Vec2-Base & $\downarrow$ & \xmark & $\uparrow$ & 46.27 & 24.72 & \textbf{30.16} & 27.17 \\
    Wav2Vec2-Base & \xmark & $\downarrow$ & $\uparrow$ & 45.27 & 22.12 & 26.84 & 24.23 \\
    Wav2Vec2-Base & $\downarrow$ & $\downarrow$ & $\uparrow$ & \textbf{48.02} & 24.95 & 30.02 & \textbf{27.23} \\
    \midrule
    Wav2Vec2-Large & $\uparrow$ & \xmark & \xmark & 75.41 & 75.37 & 75.18 & 75.24 \\
    Wav2Vec2-Large & $\uparrow$ & $\downarrow$ & \xmark & 76.57 & 76.51 & 76.39 & 76.44 \\
    Wav2Vec2-Large & $\uparrow$ & \xmark & $\downarrow$ & 77.26 & 77.37 & 76.93 & 77.03 \\
    Wav2Vec2-Large & $\uparrow$ & $\downarrow$ & $\downarrow$ & \textbf{77.53} & \textbf{78.00} & \textbf{77.04} & \textbf{77.16} \\
    \midrule
    Wav2Vec2-Large & \xmark & $\uparrow$ & \xmark & 90.26 & 90.13 & 85.01 & 87.49 \\
    Wav2Vec2-Large & $\downarrow$ & $\uparrow$ & \xmark & 91.57 & 91.89 & 86.71 & 88.83 \\
    Wav2Vec2-Large & \xmark & $\uparrow$ & $\downarrow$ & 92.19 & 92.44 & 87.75 & 89.71 \\
    Wav2Vec2-Large & $\downarrow$ & $\uparrow$ & $\downarrow$ & \textbf{93.08} & \textbf{93.03} & \textbf{89.43} & \textbf{91.00} \\
    \midrule
    Wav2Vec2-Large & \xmark & \xmark & $\uparrow$ & 47.46 & 28.64 & 28.79 & 28.71 \\
    Wav2Vec2-Large & $\downarrow$ & \xmark & $\uparrow$ & 48.76 & 23.80 & 28.76 & 26.04 \\
    Wav2Vec2-Large & \xmark & $\downarrow$ & $\uparrow$ & 49.45 & 30.27 & \textbf{29.79} & 30.02 \\
    Wav2Vec2-Large & $\downarrow$ & $\downarrow$ & $\uparrow$ & \textbf{49.93} & \textbf{34.36} & 29.66 & \textbf{31.83} \\
    \bottomrule
    \end{tabular}
    \label{tab:results}
\end{table*}

\subsection{Task-Specific Performance}

\paragraph{Spoken Dialect Identification.} For dialect identification, Wav2Vec2-Base achieves the highest F1-score (78.05\%) when using only gender as an adversarial target, representing a significant improvement over the baseline (71.80\%). With Wav2Vec2-Large, the optimal F1 performance (77.16\%) is achieved when both gender and age serve as adversarial targets, suggesting that larger models better manage multiple adversarial constraints.

\paragraph{Gender Classification.} The gender classification shows strong baseline F1 performance (87.04\% for Wav2Vec2-Base and 87.49\% for Wav2Vec2-Large). Using both dialect and age as adversarial targets yields the highest F1-scores: 88.90\% for Wav2Vec2-Base and 91.00\% for Wav2Vec2-Large, confirming consistent benefits from multi-target adversarial training.

\paragraph{Age Classification.} Age prediction demonstrates lower F1-scores overall, with baselines of 26.27\% (Wav2Vec2-Base) and 28.71\% (Wav2Vec2-Large). For both models, employing dialect and gender jointly as adversarial targets yields the best F1 performance: 27.23\% for Wav2Vec2-Base and 31.83\% for Wav2Vec2-Large, indicating a more substantial improvement for the larger model.

\subsection{Error Analysis}

\begin{figure*}
    \centering
    \includegraphics[width=1\linewidth]{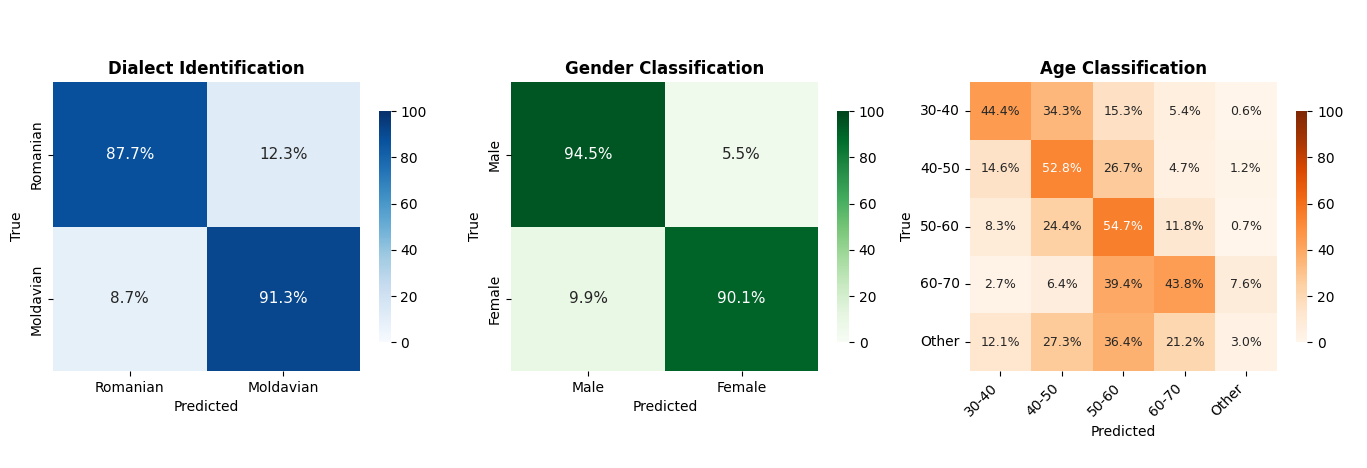}
    \caption{Confusion matrices for the best-performing models on three classification tasks using the MoRoVoc test set. Values represent classification percentages for each true-predicted class pair. (a) Spoken dialect identification using Wav2Vec2-Large with gender and age as adversarial targets. (b) Gender classification using Wav2Vec2-Large with dialect and age as adversarial targets. (c) Age classification using Wav2Vec2-Large with dialect and gender as adversarial targets.}
    \label{fig:conf_matrix}
\end{figure*}

Figure \ref{fig:conf_matrix} presents confusion matrices for the best-performing model configurations on each task. Dialect identification reveals an asymmetric misclassification: Romanian speakers are misclassified as Moldavian 12.3\% of the time, while the reverse occurs only 8.7\% of the time. This pattern likely reflects the linguistic relationship between dialects, where the Romanian language spoken in the Republic of Moldova preserves archaic Romanian features alongside Russian-influenced phonetic patterns, and standard Romanian exhibits greater internal variation, particularly among speakers from eastern regions who display transitional dialectal features.

Gender classification achieves 93.08\% accuracy with a slight male bias (94.5\% vs. 90.1\% for female speakers), potentially due to dataset imbalance and the neutralizing effect of formal speech on gender-specific prosody. The age classification performs poorly (49.93\% accuracy), with strong confusion between adjacent groups: 30-40-year-olds are correctly identified only 45\% of the time and are frequently confused with the 40-50 group (35\%). At the same time, the sparse category "Other" achieves just 10\% accuracy. This systematic pattern suggests that formal parliamentary discourse significantly attenuates age-distinctive acoustic features, such as speech rate and voice quality, as speakers adopt standardized delivery patterns regardless of age.

\subsection{Discussion}

Wav2Vec2-Large consistently outperforms Wav2Vec2-Base on all tasks, with the most substantial improvements in the F1-score for gender classification (3.50\%) and age prediction (4.60\%). The performance gap for spoken dialect identification is smaller (1.01\%), suggesting that even the smaller model can capture the linguistic features used in this task.

Our results demonstrate that multi-target adversarial training enhances performance; however, the optimal configuration depends on both the task and model size, a finding also observed in other works. This pattern indicates complex relationships between speech attributes, which our meta-learning approach effectively navigates by finding an optimal balance between task-specific discriminative power and attribute invariance.

\section{Conclusion}

This paper introduced MoRoVoc, the largest corpus of Romanian spoken dialect identification to date, comprising over 93 hours of speech and 88,192 audio samples balanced between standard Romanian and Moldavian dialects, with comprehensive gender and age annotations. We also proposed a novel multi-target adversarial training framework for fine-tuning speech models that incorporates demographic attributes as adversarial targets with coefficients dynamically adjusted through meta-learning, yielding significant performance improvements across all classification tasks: Wav2Vec2-Base achieved 78.21\% accuracy for dialect identification (a 5.20\% improvement over baseline) when using gender adversarially. At the same time, Wav2Vec2-Large reached 93.08\% accuracy for gender classification when using both dialect and age adversarially.

\section*{Acknowledgements}
This work was supported by the National University of Science and Technology POLITEHNICA Bucharest through the PubArt program.

\section{Limitations}

Despite MoRoVoc's contributions, several limitations warrant consideration. Our dataset relies exclusively on parliamentary recordings with formal speech registers, which may not generalize well to colloquial contexts. The demographic distribution is skewed (67.9\% male, 76.4\% aged 40--60), with minimal representation of younger and older speakers. This could impact the model's generalization capabilities across different groups, which may explain its modest performance on the age classification task.

Binary dialect classification (standard Romanian vs. Moldavian) also simplifies Romania's dialectal landscape, which traditionally includes five major regional variants as seen in the RoDia dataset \cite{codrut2024rodia}. Future work should diversify speech registers, improve demographic balance, and evaluate performance in various acoustic environments and model architectures. Additionally, exploring the application of our multi-target adversarial training framework to other speech recognition tasks beyond dialect identification could further validate its effectiveness.

\section{Ethical Considerations}

The MoRoVoc corpus was developed with strict adherence to ethical research standards. All audio recordings were sourced from publicly available parliamentary sessions in Romania and the Republic of Moldova, with proper documentation of sources and metadata. Speaker identities are protected in accordance with established privacy protocols, ensuring that our resources advance speech technology research while upholding legal and ethical standards for the distribution of speech data.

\bibliography{custom}

\appendix

\section{Related Work}
\label{app:related_work}

Spoken dialect identification (SDI) has gained considerable attention in recent years due to its importance in addressing linguistic diversity and improving speech technologies. Several datasets and methodologies have been developed for languages, such as Mandarin \cite{zhao2022mandi}, Arabic \citep{shon2018convolutional,alvarez2020learning,barakat2024arabic}, Spanish \citep{vasilescu2018exploring,escobar2021colombian}, Algerian \cite{lounnas2022towards}, Italian \cite{la2024speech}, Styrian \cite{yeh2019using}, Vietnamese \cite{dinh2024multi}, German \cite{dobbriner2019implementing}, and Japanese \cite{imaizumi2022end}, each exploring different aspects of dialectal speech. 

In Chinese speech processing, datasets such as KeSpeech \cite{tang2021kespeech} have showcased large-scale dialectal speech collections that include multiple subdialects of Mandarin. This dataset, with 1,542 hours of speech from over 27,000 speakers, supports various tasks, including SDI and automatic speech recognition (ASR). Similarly, Arabic dialect identification (ADI) datasets such as ADI-5 \cite{ali2017speech} and ADI-17 \cite{ali2019mgb} offer resources for dialectal speech recognition, demonstrating the potential of transfer learning and self-supervised learning models for robust dialect identification in domain shifts \cite{kulkarni-aldarmaki-2023-yet}. In contrast, Japanese SDI focuses on tightly coupled linguistic and acoustic features, as seen in studies employing multi-task learning for end-to-end SDI and ASR systems \cite{imaizumi2022end}.

For low-resource languages, initiatives to identify German dialects in Irish and Swiss have explored methods to handle limited data while emphasizing dialectal distinctions. For Irish, experiments with ECAPA-TDNN \cite{desplanques2020ecapa} and XLS-R \cite{babu2022xls} illustrated the effectiveness of combining acoustic and textual information for SDI \cite{lonergan2023towards}. Similarly, RoDia \cite{codrut2024rodia}, a pioneering dataset for identifying Romanian dialects, represents a significant step in low-resource language processing. RoDia comprises speech samples from five Romanian dialects and provides baseline models for evaluation. However, its limited scale, with only two hours of annotated speech, reflects the challenges of resource constraints.

Although RoDia set an important precedent for Romanian SDI, our newly introduced MoRoVoc dataset addresses its limitations by offering a significantly larger collection of annotated speech samples. This scale advantage underscores MoRoVoc as a crucial enhancement in the advancement of Romanian SDI research.

\section{Textual Statistics}
\label{app:text_stats}

We used the turbo version of the Whisper model \cite{radford2023robust} to generate transcriptions for the speech samples available in MoRoVoc. The RoQLlama-7b tokenizer \cite{dima2024roqllama} was then used to calculate the token statistics for these transcriptions. This process allowed us to quantify the textual characteristics of each dialect represented in our corpus.

As shown in Table \ref{tab:text_stats}, the MoRoVoc dataset includes approximately 984,000 tokens for Moldavian samples and 1,064,000 tokens for standard Romanian samples. In particular, standard Romanian samples contain an average of 24.64 tokens per sample, which is slightly higher than the Moldavian average of 21.87 tokens per sample. This slight difference in verbosity between dialects could be attributed to the varying speaking styles or discourse patterns characteristic of each region. Overall, the MoRoVoc dataset contains a substantial 88k samples across both dialects, yielding approximately 2 million tokens in total. This represents a significant textual corpus with an average of 23.19 tokens per sample, demonstrating the robustness of the dataset for computational linguistic analysis and natural language processing tasks.

\begin{table}
    \centering
    \resizebox{0.475\textwidth}{!}{
    \begin{tabular}{|l|c|c|c|}
         \toprule
    \multirow{ 2}{*}{\textbf{Dialect}} & \multirow{ 2}{*}{\textbf{\# Samples}} & \multirow{ 2}{*}{\textbf{\# Tokens}} & \textbf{Avg. Tok.} \\
    & & & \textbf{per Sample} \\
    \midrule
    Moldavian & 45k & 984k & 21.87 \\
    Standard Romanian & 43k & 1,064k & 24.64 \\
    \midrule
    \textbf{Overall} & \textbf{88k} & \textbf{2,049k} & \textbf{23.19} \\
    \bottomrule
    \end{tabular}
    }
    \caption{Textual statistics of the MoRoVoc dataset showing sample count, total token count, and average tokens per sample for each dialect.}
    \label{tab:text_stats}
\end{table}

\begin{table*}
\centering
\begin{tabular}{l|l|l|c}
    \toprule
         \textbf{Country} & \textbf{Word} & \textbf{Translation} & \textbf{TF-IDF Score} \\
         \midrule
         \multirow{10}{*}{\textbf{Romania}}& moldova & Moldova & 0.054 \\
          & parte & part & 0.047 \\
          & dumneavoastră & you (formal) & 0.046 \\
          & vedem & we see & 0.038 \\
          & domnul & the gentleman/sir & 0.038 \\
          & stat & state & 0.037 \\
          & cadrul & the framework & 0.035 \\
          & fapt & fact & 0.035 \\
          & lucru & thing/work & 0.035 \\
          & partea & the part & 0.034 \\
          \midrule
         \multirow{10}{*}{\textbf{Republic of Moldova}} & senator & senator & 0.139 \\
          & domnul & the gentleman/sir & 0.123 \\
          & punctul & the point & 0.103 \\
          & românia & Romania & 0.085 \\
          & zi & day & 0.066 \\
          & lege & law & 0.065 \\
          & domnule & sir & 0.063 \\
          & vedem & we see & 0.059 \\
          & privind & regarding & 0.058 \\
          & vot & vote & 0.058 \\
    \bottomrule
\end{tabular}
\caption{Top ten most distinctive words for standard Romanian and Moldavian dialects based on TF-IDF analysis of transcribed speech. The table includes the original words, their English translations, and corresponding TF-IDF scores indicating the relative distinctiveness of each term within its dialect.}
\label{tab:tfidf}
\end{table*}

We also performed a Term Frequency-Inverse Document Frequency (TF-IDF) \cite{leskovec2020mining} analysis of the transcriptions to identify the most distinctive terms for each dialect. The results presented in Table \ref{tab:tfidf} reveal interesting lexical preferences between the two dialects. For standard Romanian speakers, terms such as "Moldova", "part" (eng. "part"), and the formal pronoun "dumneavoastră" (eng. "you") have high distinctiveness scores. This suggests frequent references to Moldova and possibly a more formal register in parliamentary discourse.

For Moldavian speakers, terms such as "senator," "domnul" (eng. "the gentleman/sir"), and "punctul" (eng. "the point") emerge as highly distinctive. Interestingly, "românia" (eng. "Romania") appears as a characteristic term in Moldavian speech, indicating frequent references to Romania in the parliament of the Republic of Moldova. This pattern aligns with the expected topics of geopolitical discourse between these neighboring regions.

These lexical differences, while subtle, provide valuable linguistic insights into how the two dialects differ in formal parliamentary settings. The TF-IDF analysis highlights both content-based differences (references to the respective countries) and potentially stylistic variations in formal address and discourse patterns between standard Romanian and Moldavian dialects.

\section{Additional Model Results}
\label{app:additional_models}

To further validate the effectiveness of our multi-target adversarial training framework, we extended our experiments to include the HuBERT \cite{hsu2021hubert} and WavLM-Base \cite{chen2022wavlm} models. These additional experiments demonstrate the generalizability of our approach across different self-supervised speech representation architectures. Table \ref{tab:additional_results} presents the complete results for both models in all task configurations.

\begin{table*}[!ht]
    \centering
    \caption{HuBERT and WavLM-Base results on MoRoVoc. Up arrow ($\uparrow$) depicts the target optimization objective, down arrow ($\downarrow$) depicts the adversarial optimization objective(s), while \xmark \, outlines that we did not use that optimization objective.}
    \begin{tabular}{|l|wc{3.2em}|wc{3.2em}|wc{3.2em}|c|c|c|c|}
    \toprule
    \textbf{Model} & \textbf{Dialect} & \textbf{Gender} & \textbf{Age} & \textbf{Acc.} & \textbf{P} & \textbf{R} & \textbf{F1} \\
    \midrule
    HuBERT & $\uparrow$ & \xmark & \xmark & 75.87 & 76.12 & 75.44 & 75.63 \\
    HuBERT & $\uparrow$ & $\downarrow$ & \xmark & \textbf{80.34} & \textbf{80.41} & \textbf{80.15} & \textbf{80.22} \\
    HuBERT & $\uparrow$ & \xmark & $\downarrow$ & 78.92 & 78.88 & 78.71 & 78.76 \\
    HuBERT & $\uparrow$ & $\downarrow$ & $\downarrow$ & 77.15 & 77.34 & 77.08 & 77.19 \\
    \midrule
    HuBERT & \xmark & $\uparrow$ & \xmark & 90.43 & 90.76 & 85.92 & 88.27 \\
    HuBERT & $\downarrow$ & $\uparrow$ & \xmark & 91.15 & 91.34 & 87.88 & 89.58 \\
    HuBERT & \xmark & $\uparrow$ & $\downarrow$ & 90.67 & 91.02 & 86.45 & 88.68 \\
    HuBERT & $\downarrow$ & $\uparrow$ & $\downarrow$ & \textbf{91.84} & \textbf{91.67} & \textbf{88.76} & \textbf{90.19} \\
    \midrule
    HuBERT & \xmark & \xmark & $\uparrow$ & 49.18 & 28.34 & 30.12 & 29.19 \\
    HuBERT & $\downarrow$ & \xmark & $\uparrow$ & 48.65 & 26.88 & 31.45 & 28.99 \\
    HuBERT & \xmark & $\downarrow$ & $\uparrow$ & 47.93 & 25.67 & 29.34 & 27.38 \\
    HuBERT & $\downarrow$ & $\downarrow$ & $\uparrow$ & \textbf{50.34} & 27.12 & \textbf{31.88} & \textbf{29.31} \\
    \midrule
    WavLM-Base & $\uparrow$ & \xmark & \xmark & 76.54 & 77.23 & 76.12 & 76.34 \\
    WavLM-Base & $\uparrow$ & $\downarrow$ & \xmark & \textbf{81.76} & \textbf{81.88} & \textbf{81.52} & \textbf{81.67} \\
    WavLM-Base & $\uparrow$ & \xmark & $\downarrow$ & 79.88 & 79.95 & 79.65 & 79.78 \\
    WavLM-Base & $\uparrow$ & $\downarrow$ & $\downarrow$ & 78.34 & 78.67 & 78.12 & 78.28 \\
    \midrule
    WavLM-Base & \xmark & $\uparrow$ & \xmark & 91.12 & 91.45 & 86.67 & 89.00 \\
    WavLM-Base & $\downarrow$ & $\uparrow$ & \xmark & 91.88 & 92.01 & 88.34 & 90.14 \\
    WavLM-Base & \xmark & $\uparrow$ & $\downarrow$ & 91.45 & 91.78 & 87.23 & 89.45 \\
    WavLM-Base & $\downarrow$ & $\uparrow$ & $\downarrow$ & \textbf{92.56} & \textbf{92.45} & \textbf{89.45} & \textbf{90.93} \\
    \midrule
    WavLM-Base & \xmark & \xmark & $\uparrow$ & 50.12 & \textbf{29.45} & 31.23 & 30.30 \\
    WavLM-Base & $\downarrow$ & \xmark & $\uparrow$ & 49.78 & 27.56 & 32.45 & 29.82 \\
    WavLM-Base & \xmark & $\downarrow$ & $\uparrow$ & 48.89 & 26.89 & 30.78 & 28.71 \\
    WavLM-Base & $\downarrow$ & $\downarrow$ & $\uparrow$ & \textbf{51.45} & 28.34 & \textbf{32.89} & \textbf{30.44} \\
    \bottomrule
    \end{tabular}
    \label{tab:additional_results}
\end{table*}

HuBERT achieves its best performance for the spoken dialect identification task, with an F1-score of 80.22\% when using gender as the only adversarial goal, representing a 4.59\% improvement over its baseline (75.63\%). This pattern closely mirrors the results obtained with Wav2Vec2-Base, suggesting that gender-based adversarial training consistently enhances dialect discrimination across different architectures. For gender classification, HuBERT obtains an F1-score of 90.19\% when employing both dialect and age as adversarial targets, while age classification yields a modest 29.31\% F1-score under the same multi-adversarial configuration. Notably, HuBERT outperforms Wav2Vec2-Base in spoken dialect identification tasks (80.22\% vs. 78.05\% F1-score with gender as the adversarial objective), indicating its superior capability to capture dialectal acoustic variations.

WavLM-Base performs even better, obtaining the highest dialect identification F1-score of 81.67\% among all base-sized models when using gender as an adversarial goal, an improvement of 5.33\% over its baseline. This model also excels in gender classification, achieving an F1-score of 90.93\% with both dialect and age as adversarial targets, surpassing both Wav2Vec2-Base (88.90\%) and HuBERT (90.19\%). For age classification, WavLM-Base achieves 30.44\% F1-score, marginally outperforming other base models. The consistent superiority of WavLM-Base across all tasks can be attributed to its joint training on both speech and denoising objectives, which potentially provides more robust representations for handling the acoustic variations present in parliamentary speech recordings. These results reinforce our finding that multi-target adversarial training yields significant improvements regardless of the underlying self-supervised architecture, with optimal configurations varying based on the specific model's inductive biases.

\end{document}